\documentclass[11pt,a4paper]{article}
\usepackage[hyperref]{acl2021}
\usepackage{times}
\usepackage{latexsym}

\usepackage{amsmath}
\usepackage{amsfonts}
\usepackage{booktabs}

\usepackage{framed,color,xcolor}
\usepackage{multirow}
\usepackage{nicefrac}
\usepackage{threeparttable}

\usepackage{soul}
\usepackage{colortbl}
\usepackage{amsmath}
\usepackage{amsfonts}
\usepackage{amssymb}
\usepackage{tikz}
\usepackage{pgfplots}
\usepgfplotslibrary{fillbetween}
\usetikzlibrary{calc}
\usepackage{multirow, makecell, caption}
\usepackage{todonotes}
\usepackage{tipa}
\usepackage{subcaption}
\usepackage{stackengine}
\usepackage{mwe}
\usepackage{enumitem}
\aclfinalcopy

\usepackage{microtype}

\newenvironment{customlegend}[1][]{%
        \begingroup
        \csname pgfplots@init@cleared@structures\endcsname
        \pgfplotsset{#1}%
    }{%
        \csname pgfplots@createlegend\endcsname
        \endgroup
    }%
    \def\addlegendimage{\csname pgfplots@addlegendimage\endcsname}
\usepackage{tikz}
\usepackage{pgfplots}
\usepgfplotslibrary{fillbetween}
\usetikzlibrary{calc}

\definecolor{mygreen}{rgb}{0.0, 0.44, 0.0}

\newcommand{\ASS}{AS2}

\newcommand{\TANDA}{T{\sc AND}A}

\title{Joint Models for Answer Verification in Question Answering Systems}

\author{Zeyu Zhang\thanks{\hspace{.5em}Work done while the author was an intern at Amazon Alexa AI.}\hspace{.3em},  Thuy Vu, \and  Alessandro Moschitti\\
  School of Information, The University of Arizona, Tucson, AZ, USA \\
  Amazon Alexa AI, Manhattan Beach, CA, USA \\
  \texttt{zeyuzhang@email.arizona.edu, \{thuyvu, amosch\}@amazon.com}}

\date{}

\begin{document}
\maketitle
\begin{abstract}
This paper studies joint models for selecting correct answer sentences among the top $k$ provided by answer sentence selection (AS2) modules, which are  core components of retrieval-based Question Answering (QA) systems. Our work shows that a critical step to effectively exploit an answer set regards modeling the interrelated information between pair of answers. For this purpose, we build a three-way multi-classifier, which decides if an answer supports, refutes, or is neutral with respect to another one. More specifically, our neural architecture integrates a state-of-the-art AS2 model with the multi-classifier, and a joint layer connecting all components. We tested our models on WikiQA, TREC-QA, and a real-world dataset. The results show that our models obtain the new state of the art in AS2.
\end{abstract}

\section{Introduction}

Automated Question Answering (QA) research has received a renewed attention thanks to the diffusion of Virtual Assistants. 
Among the different types of methods to implement QA systems, we focus on Answer Sentence Selection (AS2) research, originated from TREC-QA track \cite{voorhees99trec}, as it proposes efficient models that are more suitable for a production setting, e.g., they are more efficient than those developed in machine reading (MR) work~\cite{DBLP:journals/corr/ChenFWB17}.

\citet{DBLP:conf/aaai/GargVM20} proposed the {\TANDA} approach based on pre-trained Transformer models, obtaining impressive improvement over the state of the art for AS2, measured on the two most used datasets, \mbox{WikiQA}~\cite{yang2015wikiqa} and TREC-QA~\cite{wang-etal-2007-jeopardy}. However, {\TANDA} was applied only to pointwise rerankers (PR), e.g., simple binary classifiers. \citet{DBLP:journals/corr/abs-2003-02349} tried to improve this model by jointly modeling all answer candidates with listwise methods, e.g., \cite{conf/cikm/Bian0YCL17}.
Unfortunately,  merging the embeddings from all candidates with standard approaches, e.g., CNN or LSTM, did not improve over {\TANDA}.

\begin{table}[t]
\small
\centering
\resizebox{\linewidth}{!}{%
\begin{tabular}{|lp{7.2cm}|}
\hline
$Claim$: & \textbf{Joe Walsh was inducted in 2001.}\\
$Ev_1$: & As a member of the Eagles, \textbf{Walsh was inducted} into the Rock and Roll Hall of Fame in 1998, and into the Vocal Group Hall of Fame \textbf{in 2001}.\\
$Ev_2$: & Joseph Fidler Walsh  (born November 20, 1947) is an American singer songwriter, composer, multi-instrumentalist and record producer.\\
$Ev_3$: & Walsh was awarded with the \textbf{Vocal Group Hall of Fame in 2001}.\\
\hline
\end{tabular}
}
\caption{A claim verification example from FEVER.}
\label{evaluator-input}
\vspace{-1em}
\end{table}

A more structured approach to building joint models over sentences can instead be observed in Fact Verification Systems, e.g., the methods developed in the FEVER challenge \cite{thorne-etal-2018-fever}. Such systems take a claim, e.g., \emph{Joe Walsh was inducted in 2001}, as input (see Tab.~\ref{evaluator-input}), and verify if it is valid, using related sentences called \emph{evidences} (typically retrieved by a search engine). For example, $Ev_1$, \emph{As a member of the Eagles, Walsh was inducted into the Rock and Roll Hall of Fame in 1998, and into the Vocal Group Hall of Fame in 2001}, and $Ev_3$, \emph{Walsh was awarded with the Vocal Group Hall of Fame in 2001}, support the veracity of the claim. In contrast, $Ev_2$ is neutral as it describes who Joe Walsh is but does not contribute to establish the induction.
We conjecture that supporting evidence for answer correctness in AS2 task can be modeled with a similar rationale. 

In this paper, we design joint models for AS2 based on the assumption that, given $q$ and a target answer candidate $t$, the other answer candidates, ($ c_1,..c_k$) can provide positive, negative, or neutral support to decide the correctness of $t$.
Our first approach exploits Fact Checking research: we adapted a state-of-the-art FEVER system, KGAT \cite{liu-etal-2020-fine}, for AS2. We defined a claim as a pair constituted of the question and one target answer, while considering all the other answers as \emph{evidences}. We re-trained and rebuilt all its embeddings for the AS2 task.

Our second method, Answer Support-based Reranker (ASR), is completely new, it is based on the representation of the pair, ($q$, $t$), generated by state-of-the-art AS2 models, concatenated with the representation of all the pairs ($t, c_i$). The latter summarizes the contribution of each $c_i$ to $t$ using a max-pooling operation. $c_i$ can be unrelated to ($q, t$) since the candidates are automatically retrieved, thus it may introduce just noise. To mitigate this problem, we use an Answer Support Classifier (ASC) to learn the relatedness between $t$ and $c_i$ by classifying their embedding, which we obtain by applying a transformer network to their concatenated text. ASC tunes the ($t, c_i$) embedding parameters according to the evidence that $c_i$ provides to $t$. Our Answer Support-based Reranker (ASR) significantly improves the state of the art, and is also simpler than our approach based on KGAT.

Our third method is an extension of ASR. It should be noted that, although ASR exploits the information from the $k$ candidates, it still produces a score for a target $t$ without knowing the scores produced for the other target answers. Thus, we jointly model the representation obtained for each target in a multi-ASR (MASR) architecture, which can then carry out a complete global reasoning over all target answers.

We experimented with our models over three datasets, WikiQA, TREC-QA, and Web-based Question Answering (WQA), where the latter is an internal dataset built on anonymized customer questions\footnote{We are making it available soon}.
The results show that: 
\begin{itemize}[leftmargin=1.2em]
\item ASR improves the best current model for AS2, i.e., {\TANDA} by $\sim$3\%, corresponding to an error reduction of 10\% in Accuracy, on both \mbox{WikiQA} and TREC-QA.
\item We also obtain a relative improvement of $\sim$3\% over {\TANDA} on WQA, confirming that ASR is a general solution to design accurate QA systems.
\item Most interestingly, MASR improves ASR by additional $2$\%, confirming the benefit of joint modeling. 
\end{itemize}
Finally, it is interesting to mention that MASR improvement is also due to the use of FEVER data for pre-fine-tuning ASC, suggesting that the fact verification inference and the answer support inference are similar.

\vspace{-.3em}
\section{Problem definition and related work}
\vspace{-.3em}
\label{sec:problem}
We consider retrieval-based QA systems, which are mainly constituted by (i) a search engine, retrieving documents related to the questions; and (ii) an AS2 model, which reranks passages/sentences extracted from the documents. The top sentence is typically used as final answer for the users.

\subsection{Answer Sentence Selection (AS2)}
The task of reranking answer-sentence candidates provided by a retrieval engine can be modeled with a classifier scoring the candidates.
Let $q$ be an element of the question set, $Q$, and ${\cal A}=\{c_1, \dots, c_n\}$ be a set of candidates for $q$, a reranker can be defined as $\mathcal{R}: Q \times \Pi({\cal A}) \rightarrow \Pi ({\cal A})$, where $\Pi({\cal A})$ is the set of all permutations of ${\cal A}$.
Previous work targeting ranking problems in the text domain has classified reranking functions into three buckets: pointwise, pairwise, and listwise methods.  \vspace{-.5em}
\paragraph{Pointwise reranking:}
This approach learns $p(q,c_i)$, which is the probability of $c_i$ correctly answering $q$, using a standard binary classification setting. The final rank is simply obtained sorting $c_i$, based on $p(q,c_i)$.
Previous work estimates $p(q,c_i)$ with neural models \cite{severyn2015learning}, also using attention mechanisms, e.g., Compare-Aggregate~\cite{DBLP:journals/corr/abs-1905-12897}, inter-weighted alignment networks~\cite{shen-etal-2017-inter}, and pre-trained Transformer models, which are the state of the art. ~\citet{DBLP:conf/aaai/GargVM20} proposed {\TANDA}, which is the current most accurate model on WikiQA and TREC-QA.
\vspace{-.5em}
\paragraph{Pairwise reranking:}
The method considers binary classifiers of the form $\chi(q,c_i,c_j)$ for determining the partial rank between $c_i$ and $c_j$, then the scoring function $p(q,c_i)$ is obtained by summing up all the contributions with respect to the target candidate $t=c_i$, e.g., $p(q,c_i) = \sum_j \chi(q,c_i,c_j)$.
There has been a large body of work preceding Transformer models, e.g., \cite{laskar-etal-2020-contextualized,tayyar-madabushi-etal-2018-integrating,conf/cikm/RaoHL16}. However, these methods are largely outperformed by the pointwise {\TANDA} model. \vspace{-.5em}
\paragraph{Listwise reranking:} This approach, e.g., \cite{conf/cikm/Bian0YCL17,cao2007learning,DBLP:journals/corr/abs-1804-05936}, aims at learning $p(q,\pi), \pi \in \Pi({\cal A})$, using the information on the entire set of candidates.~The loss function for training such networks is constituted by the contribution of all elements of its ranked items.
\subsection{Joint Modeling in QA}
MR is a popular QA task that identifies an answer string in a paragraph or a text of limited size for a question.
Its application to retrieval scenario has also been studied~\cite{DBLP:journals/corr/ChenFWB17,DBLP:journals/corr/abs-1906-04618,kratzwald-feuerriegel-2018-adaptive}. Their joint modeling aspect regards combining sentences from the same paragraphs.
However, the large volume of retrieved content makes their use not practical yet.

\citet{jin2020ranking}  use the relation between candidates in Multi-task learning approach for AS2. They do not exploit transformer models, thus their results are rather below the state of the art.
\citet{DBLP:journals/corr/abs-2003-02349} designed several joint models, which improved early neural networks based on CNN and LSTM for AS2, but failed to improve the state of the art using pre-trained Transformer models.

In contrast to the work above, our modeling is driven by an answer support strategy, where the pieces of information are taken from different documents. This makes our model even more unique; it allows us to design innovative joint models, which are still not designed in any MR systems.

\subsection{Fact Verification for Question Answering}

Fact verification has become a social need given the massive amount of information generated daily.
The problem is, therefore, becoming increasingly important in NLP context~\cite{DBLP:journals/corr/abs-1803-03178}.
In QA, answer verification is directly relevant due to its nature of content delivery~\cite{mihaylova-etal-2019-semeval}.
This was explored in MR setting by \citet{wang-etal-2018-multi-passage}.
Recently, \citet{DBLP:journals/corr/abs-2010-04970} proposed to use potential candidates to extract evidence in ranking.
~\citet{zhang-etal-2020-answerfact} proposed to fact check for product questions using additional associated evidence sentences.

The latter are retrieved based on similarity scores computed with both TF-IDF and  sentence-embeddings from pre-trained BERT models.
While the process is technically sound, the retrieval of evidence is an expensive process, which is prohibitive to scale in production.
We instead address this problem by leveraging the top answer candidates.

\section{Baseline Models for AS2}

In this section, we describe our baseline models, which are constituted by pointwise, pairwise, and listwise strategies.

\subsection{Pointwise Models}
\label{sec:baseline}

One simple and effective method to build an answer selector is to use a pre-trained Transformer model, adding a simple classification layer to it, and fine-tuning the model on the AS2 task.  Specifically, $q =$ Tok$^q_1$,...,Tok$^q_N$   and $c=$Tok$^c_1$,...,Tok$^c_M$ are encoded in the input of the Transformer by delimiting them using three tags: [CLS], [SEP] and [EOS], inserted at the beginning, as separator, and at the end, respectively. This input is encoded as three embeddings based on tokens, segments and their positions, which are fed as input to several layers (up to 24). Each of them contains sublayers for multi-head attention, normalization and feed forward processing. The result of this transformation is an embedding, $\mathbf{E}$, representing $(q,c)$, which models the dependencies between words and segments of the two sentences.

For the downstream task, $\mathbf{E}$ is fed (after applying a non-linearity function) to a fully connected layer having weights: $W$ and $B$. The output layer can be used to implement the task function. For example, a softmax can be used to model the probability of the question/candidate pair classification, as
$p(q,c) = softmax(W \times tanh(E(q,c)) + B)$. 

We can train this model with log cross-entropy loss: $\mathcal{L}=-\sum_{l \in \{0,1\}} y_l \times log(\hat{y}_l)$ on pairs of texts, where $y_l$ is the correct and incorrect answer label, $\hat{y}_1 = p(q,c)$, and $\hat{y}_0 = 1-p(q,c)$. 
Training the Transformer from scratch requires a large amount of labeled data, but it can be pre-trained using a masked language model, and the next sentence prediction tasks, for which labels can be automatically generated. Several methods for pre-training Transformer-based language models have been proposed, e.g., BERT~\cite{DBLP:journals/corr/abs-1810-04805}, RoBERTa~\cite{DBLP:journals/corr/abs-1907-11692}, XLNet~\cite{DBLP:journals/corr/abs-1906-08237}, AlBERT~\cite{lan2019albert}.

\subsection{Our joint model baselines}

To better show the potential of our approach and the complexity of the task, we designed three joint model baselines based on: (i) a multiclassifier approach (a listwise method), and (ii) a pairwise joint model operating over $k+1$ candidates, and our adaptation of KGAT model (a pairwise method).

\paragraph{Joint Model Multi-classifier}
The first baseline is also a Transformer-based architecture: we concatenate the question with the top $k+1$ answer candidates, i.e., $\left(q [SEP] c_1 [SEP] c_2 \dots [SEP] c_{k+1}\right)$, and provide this input to the same Transformer model used for pointwise reranking.  We use the final hidden vector $E$ corresponding to the first input token $[CLS]$ generated by the Transformer, and a classification layer with weights $W \in R^{{(k+1)} \times |E|}$, and train the model using a standard cross-entropy classification loss: $y \times log(softmax(EW^T))$, where $y$ is a one-hot vector representing labels for the $k+1$ candidates, i.e., $\left | y\right | =  k+1$. We use transformer models such as RoBERTa-Base (or Large), and fine-tuned them with the TANDA approach, i.e., using ASNQ~\cite{DBLP:conf/aaai/GargVM20} as first fine-tuning. The scores for the candidate answers are calculated as $\big(p(c_1),..,p(c_{k+1})\big)=softmax(EW^T)$. Then, we rerank $c_i$ according their probability.
\paragraph{Joint Model Pairwise}
To obtain our second baseline, we concatenate the question with each $c_i$ to constitute the $(q, c_i)$ pairs, which are input to the Transformer, and we use the first input token $[CLS]$ as the representation of each $(q, c_i)$ pair. Then, we concatenate the embedding of the pair containing the target candidate, $(q, t)$ with the embedding of all the other candidates' $[CLS]$. $(q, t)$ is always in the first position. We train the model using a standard classification loss. At classification time, we select one target candidate at a time, and set it in the first position, followed by all the others. We classify all $k+1$ candidates and use their score for reranking them. It should be noted that to qualify for a pairwise approach, Joint Model Pairwise should use a ranking loss. However, we always use standard cross-entropy loss as it is more efficient and the difference is performance is negligible.
\paragraph{Joint Model with KGAT}
\label{sec:baseline_kgat}

\citet{liu-etal-2020-fine} presented an interesting model, Kernel Graph Attention Network (KGAT), for fact verification: given a claimed fact $f$, and a set of \emph{evidences} $Ev=\{ev_1, ev_2, \dots, ev_m\}$, their model carries out joint reasoning over $Ev$, e.g., aggregating information to estimate the probability of $f$ to be true or false, $p(y|f,Ev)$, where $y\in$\{true, false\}. 

The approach is based on a fully connected graph, $G$, whose nodes are the $n_i=(f, ev_i)$ pairs, and $p(y |f, Ev) = p(y | f, ev_i, Ev) p(ev_i | f, Ev)$, where $p(y | f, ev_i, Ev)=p(y|n_i, G)$ is the label probability in each node $i$ conditioned on the whole graph, and $p(ev_i | f, Ev) = p(n_i | G)$ is the probability of selecting the most informative evidence. KGAT uses an edge kernel to perform a hierarchical attention mechanism, which propagates information between nodes and aggregate evidences.

We built a KGAT model for AS2 as follows: we replace (i) $ev_i$ with the set of candidate answers $c_i$, and (ii) the claim $f$ with the question and a target answer pair, $(q,t)$. KGAT constructs the evidence graph $G$ by using each claim-evidence pair as a node, which, in our case, is $((q,t),c_i)$, and connects all node pairs with edges, making it a fully-connected evidence graph. This way, sentence and token attention operate over the triplets, $(q,t,c_i)$, establishing semantic links, which can help to support or undermine the correctness of $t$. 

The original KGAT aggregates all the pieces of information we built, based on their relevance, to determine the probability of $t$. As we use AS2 data, the probability will be about the correctness of $t$. More in detail, we initialize the node representation using the contextual embeddings obtained with two TANDA-RoBERTa-base models \footnote{https://github.com/alexa/wqa\_tanda}: the first produces the embedding of $(q,t)$, while the second outputs  the embedding of $(q,c_i)$. Then, we apply a max-pooling operation on these two to get the final node representation. The rest of the architecture is identical to the original KGAT. 
Finally, at test time, we select one $c_i$ at a time, as the target $t$, and compute its probability, which ranks $c_i$.

\section{Joint Answer Support Models for AS2}

We propose the Answer Support Reranker (ASR), which uses an answer pair classifier to provide evidence to a target answer $t$. 
Given a question $q$, and a subset of its top-$k$+1 ranked answer candidates, ${\cal A}$ (reranked by an AS2 model), we build a function, $\sigma:Q \times C \times C^k\rightarrow \mathbb{R}$ such that $\sigma(q, t, {\cal A} \setminus \{t\})$ provides the probability of $t$ to be correct, where $C$ is the set of sentence-candidates.
We also design a multi-classifier MASR, which combines $k$ ASR models, one for each different target answer.

\begin{figure}[t]
  \begin{subfigure}[b]{0.48\linewidth}
      \centering
      \hspace{-5mm}
      \vspace{-1pt}
      \includegraphics[width=1.11\linewidth]{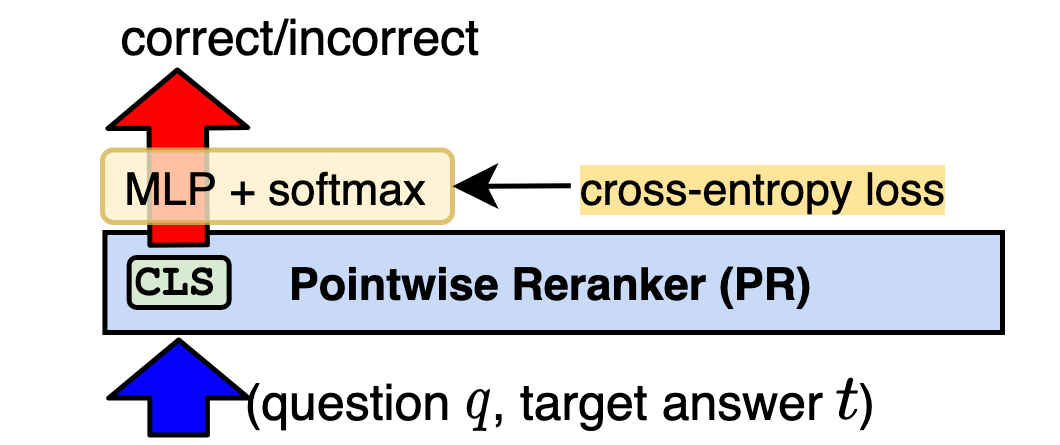}
      \caption{Baseline Reranker using Transformers.}
      \label{fig-brr}
  \end{subfigure}
  \hspace{1pt}
  \begin{subfigure}[b]{.48\linewidth}
    \centering
      \hspace{-5mm}
      \includegraphics[width=1.1\linewidth]{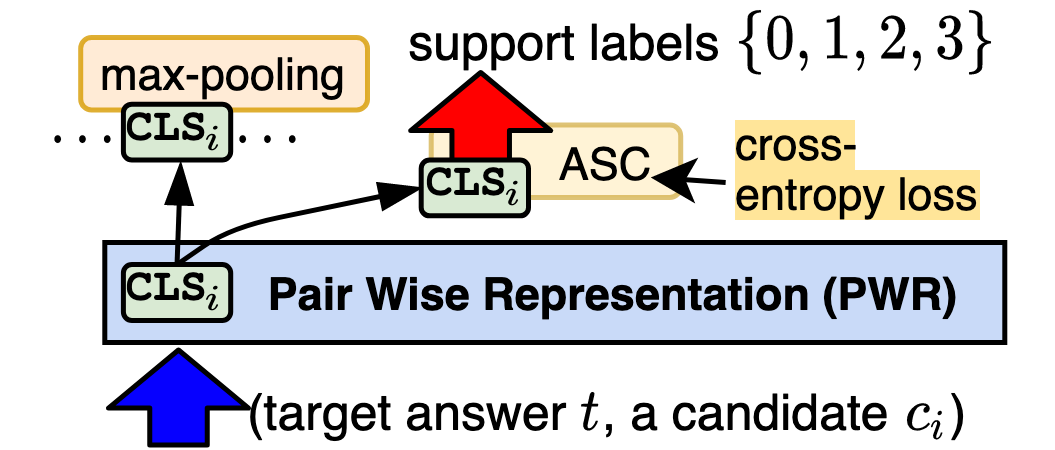}
    \caption{PairWise Representation using Transformers.}
    \label{fig-acs}
  \end{subfigure}
  \begin{subfigure}[b]{\linewidth}
      \centering
      \includegraphics[width=0.7\linewidth]{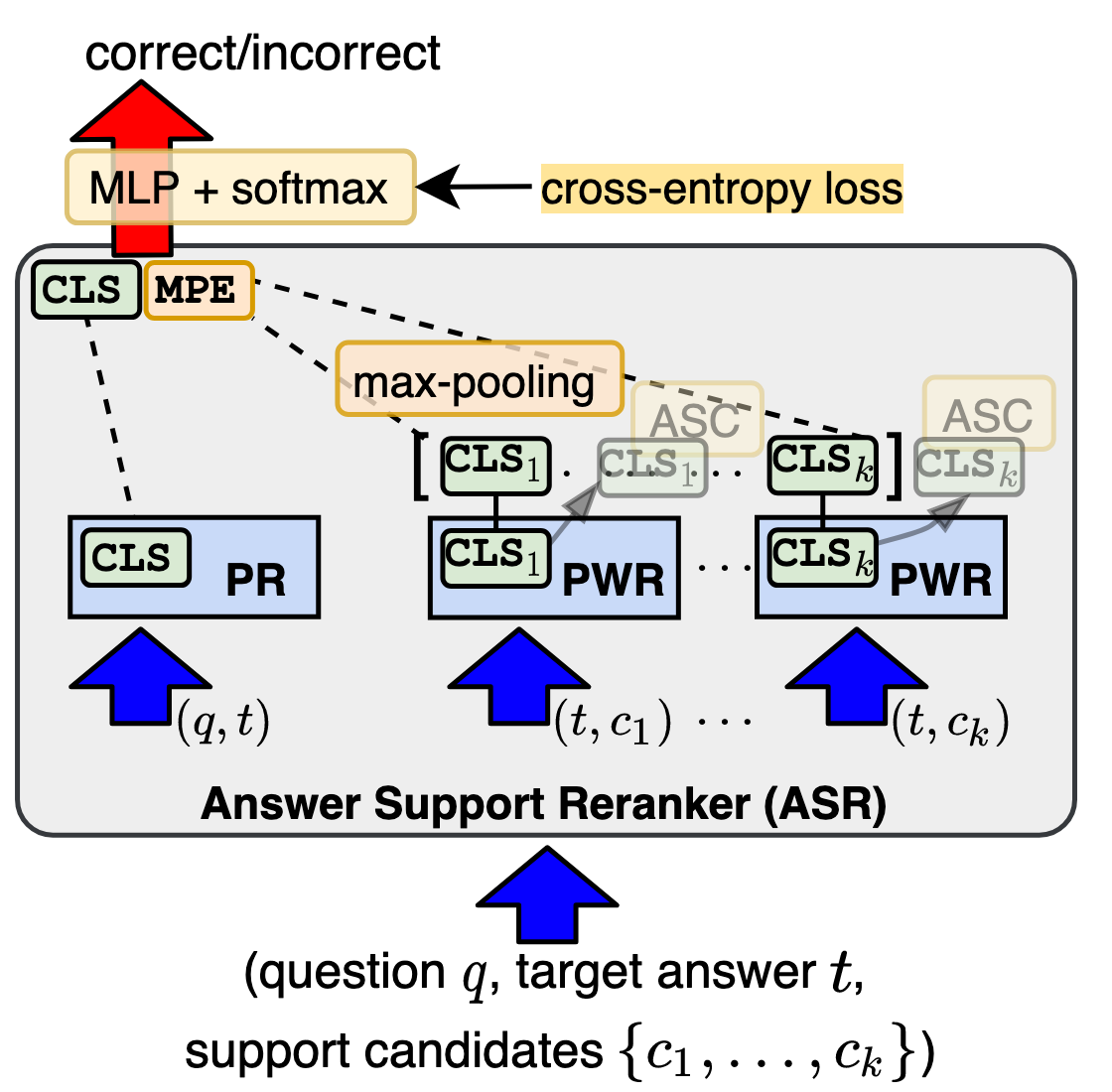}
      \caption{Answer Support Reranker}
      \label{fig-asr}
  \end{subfigure}
  \begin{subfigure}[b]{\linewidth}
    \centering
      \includegraphics[width=0.7\linewidth]{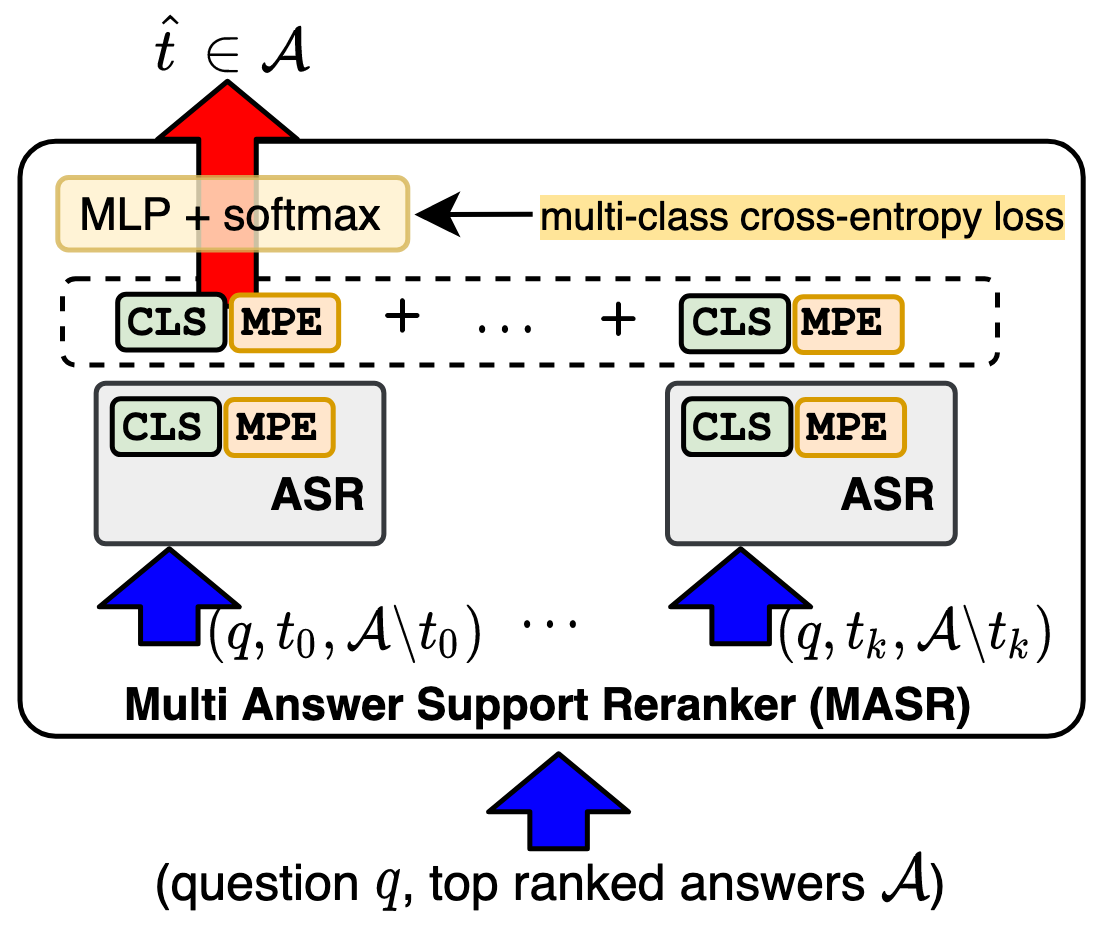}
    \caption{Multi Answer Support Reranker}
    \label{fig-masr}
  \end{subfigure}
  \caption{Multi-Answer Support Reranker and its building blocks.}
  \label{fig:sizeboxed-does-well}
\end{figure}

\subsection{Answer Support-based Reranker (ASR)}
\label{sec:new_model}

We developed ASR architecture described in Figure~\ref{fig-asr}.
This consists of three main components:
\begin{enumerate}[leftmargin=1.2em]

\item  a \emph{Pointwise Reranker} (PR), which provides the embedding of the input $(q,t)$, described in Figure~\ref{fig-brr}. This is essentially the state-of-the-art AS2 model based on the {\TANDA} approach applied to RoBERTa pre-trained transformer.
\item  To reduce the noise that may be introduced by irrelevant $c_i$, we use the \emph{Answer Support Classifier} (ASC), which classifies each $(t,c_i)$ in one of the following four classes: 
\begin{itemize}
\setlength\itemsep{0em}
\item [\textbf{0}]: $t$ and $c_i$ are both correct, 
\item [\textbf{1}]: $t$ is correct while $c_i$ is not,
\item [\textbf{2}]: vice versa, and 
\item [\textbf{3}]: both incorrect.
\end{itemize}
 This multi-classifier, described in Figure~\ref{fig-acs}, is built on top a RoBERTa Transformer, which produced a PairWise Representation (PWR). ASC is trained end-to-end with the rest of the network in a multi-task learning fashion, using its specific cross-entropy loss, computed with the labels above. 
\item  The ASR (see Figure~\ref{fig-asr}) uses the joint representation of $(q,t)$ with $(t,c_i)$, $i=1,..,k$, where $t$ and $c_i$ are the top-candidates reranked by PR. The $k$ representations are summarized by applying a max-pooling operation, which will aggregate all the supporting or not supporting properties of the candidates with respect to the target answer. The concatenation of the PR embedding with the max-pooling embedding is given as input to the final classification layer, which scores $t$ with respect to $q$, also using the information from the other candidates. For training and testing, we select a $t$ from the $k+1$ candidates of $q$ at a time, and compute its score. This way, we can rerank all the $k+1$ candidates with their scores.
\end{enumerate}

\vspace{-.5em}
\paragraph{Implementation details:} ASR is a PR that also exploits the relation between $t$ and ${\cal A} \setminus \{t\}$. We use RoBERTa to generate the $[CLS] \in \mathbb{R}^{d}$ embedding of $(q, t) = E_t$. We denote with $\hat E_j$ the $[CLS]$ output by another RoBERTa Transformer applied to answer pairs, i.e., $(t, c_j)$. Then, we concatenate $E_t$ to the max-pooling tensor from $\hat E_1,..,\hat E_k$:\vspace{-.9em}\\
\begin{equation}
\label{concatenation}
V=[E_t : \text{Maxpool}([\hat E_1,..,\hat E_k])],
\vspace{-.2em}
\end{equation}
where $V \in \mathbb{R}^{2d}$ is the final representation of the target answer $t$. Then, we use a standard feedforward network to implement a binary classification layer: $p(y_i | q, t, C_k)=softmax(VW^{T}+B)$, where $W \in \mathbb{R}^{2 \times 2d}$ and $B$ are parameters to transform the representation of the target answer $t$ from dimension $2d$ to dimension $2$, which represents correct or incorrect labels.

\vspace{-.5em}
\paragraph{ASC labels} There can be different interpretations when attempting to define labels for answer pairs. An alternative to the definition illustrated above is to use the following FEVER compatible encoding:  
\begin{itemize}
\setlength\itemsep{0em}
\item [\textbf{0}]: $t$ is correct, while  $c_i$ can be any value. This models $t$ correctness, as also an incorrect $c_i$ may provide important context (corresponding to FEVER Support label);
\item [\textbf{1}]: $t$ is incorrect and $c_i$ correct. This models $t$ incorrectness, since $c_i$ can provide evidence that $t$ is not similar to a correct answer (corresponding to FEVER Refutal label); and
\item [\textbf{2}]: both are incorrect. In this case, nothing can be told (corresponding to FEVER Neutral label).
\end{itemize}

\subsection{\mbox{Multi-Answer Support Reranker (MASR)}}
ASR still selects answers with a pointwise approach\footnote{Again, using ranking loss did not provide a significant improvment.}. This means that we can improve it by building a listwise model that selects the best answer for each question using the information from all target answers.
In particular, the architecture of MASR shown in Figure~\ref{fig-masr} is made up of two parts: (i) a list of {ASR} containing $k+1$ ASR blocks, in which each ASR block provides the representation of a target answer $t$. (ii) A final multiclassifier and a softmax function, which scores each $t$ from $k+1$ embedding concatenation and selects the one with highest score. For training and testing, we select the $t$ from the $k+1$ candidates of $q$ based on a softmax output at a time.

\begin{table}[t]
\centering
\resizebox{\linewidth}{!}{%
\begin{tabular}{|r|r|r|r|r|r|r|r|r|r|} 
\hline
\multirow{2}{*}{Dataset} & \multicolumn{3}{c|}{Train}                                                       & \multicolumn{3}{c|}{Dev}                                                         & \multicolumn{3}{c|}{Test}                                                         \\ 
\cline{2-10}
                         & \multicolumn{1}{c|}{\#Q} & \multicolumn{1}{c|}{\#A+} & \multicolumn{1}{c|}{\#A-} & \multicolumn{1}{c|}{\#Q} & \multicolumn{1}{c|}{\#A+} & \multicolumn{1}{c|}{\#A-} & \multicolumn{1}{c|}{\#Q} & \multicolumn{1}{c|}{\#A+} & \multicolumn{1}{c|}{\#A-}  \\ 
\hline
WikiQA                   & 873                      & 1,040                     & 7,632                     & 121                      & 140                       & 990                       & 237                      & 293                       & 2,058                      \\ 
\hline
TREC-QA                  & 1,229                    & 6,403                     & 47,014                    & 65                       & 205                       & 912                       & 68                       & 248                       & 1,194                      \\ 
\hline
WQA                     & 5,000                    & 42,962                    & 163,289                   & 905                      & 8,179                     & 28,096                    & 1,000                    & 8,256                     & 30,123                     \\ 
\hline
\end{tabular}
}
\caption{AS2 dataset statistics}
\label{table:alldatasets}
\end{table}

\paragraph{Implementation details:} The goal of MASR is to measure the relation between $k+1$ target answers, $t_0$, .., $t_k$. The representation of each target answer is the embedding $V\in \mathbb{R}^{2d}$ from Equation~\ref{concatenation} in ASR. Then, we concatenate the hidden vectors of $k+1$ target answers to form a matrix  $V_{(q, k+1)}\in \mathbb{R}^{(k+1)\times{2d}}$. We use this matrix and a classification  layer  weights $W\in \mathbb{R}^{2d}$, and compute a standard multi-class classification loss:
\begin{equation}
\mathcal{L}_{MASR} = y\times log(softmax(V_{(q, k+1)}W^{T}),
\end{equation}
where $y$ is a one-hot-vector, and $|y|=|k+1|$.

\section{Experiments}
\label{sec:experiments}
In these experiments, we compare our models: KGAT, ASR and MASR with pointwise models, which are the state of the art for AS2. We also compare them with our joint model baselines (pairwise and listwise). Finally, we provide an error analysis. 

\subsection{Datasets}
\label{sec:dataset}
We used two most popular AS2 datasets, and one real world application dataset we built to test the generality of our approach.
\paragraph{WikiQA}\hspace{-1em} is a QA dataset~\cite{yang2015wikiqa} containing a sample of questions and answer-sentence candidates from Bing query logs over Wikipedia. The answers are manually labeled. 
We follow the most used setting: training with all the questions that have at least one correct answer, and validating and testing with all the questions having at least one correct and one incorrect answer.
\paragraph{TREC-QA}\hspace{-1.2em} is another popular QA benchmark by~\citet{wang-etal-2007-jeopardy}. We use the same splits of the original data, following the common setting of previous work, e.g., \cite{DBLP:conf/aaai/GargVM20}.

\paragraph{WQA} The Web-based Question Answering is a dataset built by Alexa AI as part of the effort to improve understanding and benchmarking in QA systems.
The creation process includes the following steps: (i) given a set of questions we collected from the web, a search engine is used to retrieve up to 1,000 web pages from an index containing hundreds of millions pages.
(ii) From the set of retrieved documents, all candidate sentences are extracted and ranked using {\ASS} models from~\cite{DBLP:conf/aaai/GargVM20}.
Finally, (iii) top candidates for each question are manually assessed as correct or incorrect by human judges.
This allowed us to obtain a richer variety of answers from multiple sources with a higher average number of answers.

Table~\ref{table:alldatasets} reports the corpus statistics of WikiQA, TREC-QA, and WQA\footnote{The public version of WQA will be released in the short-term future. Please search for a publication with title \emph{WQA: A Dataset for Web-based Question Answering Tasks} on arXiv.org.}. 

\paragraph{FEVER}\hspace{-.9em} is a large-scale public corpus, proposed by~\citet{thorne-etal-2018-fever} for fact verification task, consisting of 185,455 annotated claims from 5,416,537 documents from the Wikipedia dump in June 2017. All claims are labelled as Supported, Refuted or Not Enough Info by annotators. Table~\ref{table:fever} shows the statistics of the dataset, which remains the same as in~\cite{thorne-etal-2018-fact}.

\begin{table}[t]
\small
\centering
\resizebox{0.8\linewidth}{!}{%
\begin{tabular}{|r|r|r|r|} 
\hline
Data split & \multicolumn{1}{c|}{Supported} & \multicolumn{1}{c|}{Refuted} & \multicolumn{1}{c|}{Not Enough Info}  \\ 
\hline
Train      & 80,035                 & 29,775                         & 35,639                      \\ 
\hline
Dev        & 6,666                   & 6,666                           & 6,666                        \\ 
\hline
Test       & 6,666                   & 6,666                           & 6,666                       \\
\hline
\end{tabular}
}
\arrayrulecolor{black}
\caption{FEVER dataset statistics}
\label{table:fever}
\vspace{-.5em}
\end{table}

\begin{table*}
\centering
\resizebox{.85\linewidth}{!}{
\begin{tabular}{|c|r|r|r|r|r|r|r|r|r|} 
\hline
RoBERTa Base                                                & \multicolumn{3}{c|}{WikiQA}                                                    & \multicolumn{3}{c|}{TREC-QA}                                                    & \multicolumn{3}{c|}{WQA}                                                       \\ 
\hline
                                                            	& \multicolumn{1}{l|}{P@1} & \multicolumn{1}{l|}{MAP} & \multicolumn{1}{l|}{MRR} & \multicolumn{1}{l|}{P@1} & \multicolumn{1}{l|}{MAP} & \multicolumn{1}{l|}{MRR} & \multicolumn{1}{l|}{P@1} & \multicolumn{1}{l|}{MAP} & \multicolumn{1}{l|}{MRR}  \\ 
\hline
Reranker by Garg et al., 2020                                           &            --              & 0.8890                   & 0.9010                   & --    & 0.9140                   & 0.9520                   &           --               &            --              &                --           \\ 
\hline
Our Reranker                                                    & 0.8189$^\dagger$                   & 0.8860                   & 0.8983                   & 0.9118                   & 0.9043                   & 0.9498                   &           $0^\dagger$               &          $0$                &      $0$                     \\ 
\hline
Joint Model Mu\textbf{}lti-classifier ($k$=5) & 0.7819$^\dagger$                   & 0.8542                   & 0.8684                   & 0.8971                   & 0.9052                   & 0.9424                   & -2.29$^\dagger$\%                   & -1.00\%                  & -1.23\%                    \\ 
\hline
Joint Model Pairwise ($k$=3)    & 0.8272$^\dagger$                   & 0.8927                   & 0.9045                   & 0.9559                   & 0.9196                   & 0.9743                   & 2.67$^\dagger$\%                   & 0.39\%                   & 1.39\%                   \\ 
\hline
KGAT ($k$=2)                                            & \textbf{0.8436}                   & 0.8991                   & 0.9120                   & 0.9412                   & 0.9155                   & 0.9645                   & 2.10\%                  & 0.39\%                  & 0.93\%                   \\ 
\hline
ASR ($k$=3)                                              & $^\dagger$\textbf{0.8436}              & \textbf{0.9014}                    & \textbf{0.9123}                   & \textbf{0.9706}                   & \textbf{0.9257}                   & \textbf{0.9816}                   & $^\dagger${2.86}\%                   & {0.86}\%                   & {1.39}\%                   \\
\hline
MASR ($k$=3)                                              &{0.8230}                    & {0.8891}                    & {0.9017}                   & {0.9265}                   & {0.9200}                   & {0.9632}                   & {$^\dagger$3.82}\%                   & {0.70}\%                   & {1.67}\%                   \\
\hline
MASR-F ($k$=3)                                              & {0.8272}                    & {0.8918}                    & {0.9031}                   & {0.9412}                   & {0.9222}                   & {0.9706}                   & {2.67}\%                   & {0.55}\%                   & {1.47}\%                   \\
\hline
MASR-FP ($k$=3)                                              & \textbf{$^\dagger$0.8436}                    & {0.8998}                    & {0.9113}                   & {0.9559}                   & {0.9191}                   & {0.9743}                   & \textbf{$^\dagger$4.96}\%                   & \textbf{0.94}\%                   & \textbf{2.43}\%                   \\
\hline

\end{tabular}
}
\arrayrulecolor{black}
\caption{Results on WikiQA, TREC-QA and WQA, using RoBERTa base Transformer. The WQA results are reported in terms of difference with Our Reranker baseline (for company policy). $^\dagger$ is used to indicate that the difference in P@1 between ASR and the other marked systems is statistically significant at 95\%.
}
\label{table:resultsForall_base}
\end{table*}

\subsection{Training and testing details}
\paragraph{Metrics} 
The performance of QA systems is typically measured with Accuracy in providing correct answers, i.e., the percentage of correct responses. This is also referred to Precision-at-1 (P@1) in the context of reranking, while standard Precision and Recall are not essential in our case as we assume the system does not abstain from providing answers. 
We also use Mean Average Precision (MAP) and Mean Reciprocal Rank (MRR) evaluated on the test set, using the entire set of candidates for each question (this varies according to the dataset), to have a direct comparison with the state of the art.

\paragraph{Models} We use the pre-trained RoBERTa-Base (12 layer) and RoBERTa-Large-MNLI (24 layer) models, which were released as checkpoints for use in downstream tasks\footnote{https://github.com/pytorch/fairseq}. 

\paragraph{Reranker training} 
We adopt Adam optimizer \cite{Kingma2014AdamAM} with a learning rate of 2e-5 for the transfer step on the ASNQ dataset~\cite{DBLP:conf/aaai/GargVM20}, and a learning rate of 1e-6 for the adapt step on the target dataset. 
We apply early stopping on the development set of the target corpus for both fine-tuning steps based on the highest MAP score. We set the max number of epochs equal to 3 and 9 for the adapt and transfer steps, respectively. We set the maximum sequence length for RoBERTa to 128 tokens. 



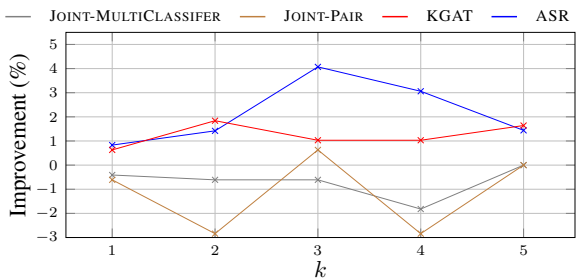
\begin{figure}[t]
\centering
\vspace{-.5em}
\resizebox{\linewidth}{!}{
\begin{tikzpicture}
        \begin{customlegend}
        	[legend columns=4,legend style={align=center,column sep=2ex, draw=none, fill=none},
        	legend entries={\textsc{Joint-MultiClassifer},\textsc{Joint-Pair},\textsc{KGAT},\textsc{ASR}}]
        \addlegendimage{gray, line width=1pt}
        \addlegendimage{brown, line width=1pt}
        \addlegendimage{red, line width=1pt}
        \addlegendimage{blue, line width=1pt}
        \end{customlegend}
\end{tikzpicture}
}

\resizebox{\linewidth}{!}{
\begin{tikzpicture}
	\begin{axis}[
	        	name=plotTest,
	        	scale=1.0,
		xlabel={$k$},
		ylabel={Improvement (\%)},
		grid=both,
		xtick={1,2,3,4,5},
		xmax=5.5,
		ytick={-3.00, -2.00, -1.00, 0.00, 1.00, 2.00, 3.00, 4.00, 5.00, 6.00},
		ymin=-3.00, 
		ymax=5.50,
		ylabel near ticks,
		xlabel near ticks,
		ylabel style={yshift=-1ex},
		tick label style={font=\scriptsize},
		xlabel style={yshift=1ex},
		anchor=north,
		width=10cm,height=5cm
		]
	\addplot [color=blue, mark=x]
	    table[x=k, y=pavm,] {data/iqad-dev-random-reranker_all.txt};
	\addplot [color=red, mark=x]
	    table[x=k, y=kgat,] {data/iqad-dev-random-reranker_all.txt};
	\addplot [color=gray, mark=x]
	    table[x=k, y=multi,] {data/iqad-dev-random-reranker_all.txt};
	\addplot [color=brown, mark=x]
	    table[x=k, y=pair,] {data/iqad-dev-random-reranker_all.txt};
	\end{axis}
\end{tikzpicture}
}
\vspace{-1.7em}
\caption{Impact of $k$ on the WQA dev.~set}
\label{fig:iqad_dev}
\vspace{-.2em}
\end{figure}

\paragraph{KGAT and ASR training} 
Again, we use the Adam optimizer with a learning rate of 2e-6 for training the ASR model on the target dataset. We utilize 1 Tesla V100 GPU with 32GB memory and a training batch size of eight.
We set the maximum sequence length for RoBERTa Base/Large to 130 tokens and the number of training epochs to 20. The other training configurations are the same of the original KGAT model from \cite{liu-etal-2020-fine}.
We use two transformer models for ASR: a RoBERTa Base/Large for PR, and one for ASC. We set the maximum sequence length for RoBERTa to 128 tokens and the number of epochs to 20.

\paragraph{MASR training} 
We use the same configuration of the ASR training, including the optimizer type, learning rate, the number of epochs, GPU type, maximum sequence length, etc. Additionally, we design two different models MASR-F, using an ASC classifier targeting the FEVER labels, and MASR-FP, which initializes ASC with the data from FEVER. This is possible as the labels are compatible.

\subsection{Choosing the best $k$}
The selection of the hyper-parameter $k$, i.e., the number of candidates to consider for supporting a target answer is rather tricky. Indeed, the standard validation set is typically used for tuning PR. This means that the candidates PR moves to the top $k+1$ positions are \emph{optimistically} accurate. Thus, when selecting also the optimal $k$ on the same validation set, there is high risk to overfit the model. 

We solved this problem by running a PR version not heavily optimized on the dev.~set, i.e., we randomly choose a checkpoint after the standard three epochs of fine-tuning of RoBERTa transformer. Additionally, we tuned $k$ only using the WQA dev.~set, which contains $\sim$36,000 Q/A pairs. WikiQA and TREC-QA dev.~sets are too small to be used (121 and 65 questions, respectively). Fig.~\ref{fig:iqad_dev} plots the improvement of four different models, Joint Model Multi-classifier, Joint Model Pairwise, KGAT, and ASR, when using different $k$ values. Their best results are reached for 5, 3, 2, and 3, respectively.  We note that the most reliable curve shape (convex) is the one of ASR and Joint Model Pairwise. 

\subsection{Comparative Results}

Table~\ref{table:resultsForall_base} reports the P@1, MAP and MRR of the rerankers, and different answer supporting models on WikiQA, TREC-QA and WQA datasets. As WQA is an internal dataset, we only report the improvement over PR in the tables. All models use RoBERTa-Base pre-trained checkpoint and start from the same set of $k$ candidates reranked by PR (state-of-the-art model).
The table shows that:
\begin{itemize}[leftmargin=1.2em]
\item PR replicates the MAP and MRR of the state-of-the-art reranker by \citet{DBLP:conf/aaai/GargVM20} on \mbox{WikiQA}.
\item Joint Model Multi-classifier performs lower than PR for all measures and all datasets. This is in line with the findings of \citet{DBLP:journals/corr/abs-2003-02349}, who also did not obtain improvement when jointly used all the candidates altogether in a representation.
\item Joint Model Pairwise differs from ASR as it concatenates the embeddings of the $(q,c_i)$, instead of using max-pooling, and does not use any Answer Support Classifier (ASC). Still, it exploits the idea of aggregating the information of all pairs $(q,c_i)$ with respect to a target answer $t$, which proves to be effective, as the model improves on PR over all measures and datasets.
\item Our KGAT version for AS2 also improves PR over all datasets and almost all measures, confirming that the idea of using candidates as support of the target answer is generally valid. However, it is not superior to Joint Model Pairwise.
\item ASR achieves the highest performance among all models (but MASR-FP on WQA), all datasets, and all measures. For example, it outperforms PR by almost 3 absolute percent points in P@1 on WikiQA, and by almost ~6 points on TREC from 91.18\% to 97.06\%, which corresponds to an error reduction of 60\%.
\item MASR and MASR-F do not achieve better performance than Joint Model Pairwise on \mbox{WikiQA} and TREC, although MASR outperforms all baselines and even ASR on WQA. This suggests that the significantly higher number of parameters of MASR cannot be trained on small corpora, while WQA has a sufficient number of examples.
\item MASR-FP exploiting FEVER for the initialization of ASC performs better than MASR and MASR-F on WikiQA and TREC. Interestingly, it significantly  outperforms ASR by 2\% on WQA. This confirms the potential of the model when enough training data is available.

\item We perform randomization test \cite{DBLP:journals/corr/cs-CL-0008005} to verify if the models significantly differ in terms of prediction outcome. We use 100,000 trials for each calculation. The results confirm the statistically significant difference of ASR and MASR over the baselines, i.e., Our Reranker, Joint Model Multi-classifier, and Joint Model Pairwise, with p $<$ 0.05 on both WikiQA and WQA.

\end{itemize}

\subsection{Official State of the art} As the state of the art for AS2 is obtained using RoBERTa Large, we trained KGAT and ASR using this pre-trained language model. Table~\ref{table:resultsForall_large} also reports the comparison with PR, which is the official state of the art. Again, our PR replicates the results of \citet{DBLP:conf/aaai/GargVM20}, obtaining slightly lower performance on WikiQA but higher on TREC-QA. KGAT performs lower than PR on both datasets. 

ASR establishes the new state of the art on \mbox{WikiQA} with an MAP of 92.80 vs.~92.00. The P@1 also significantly improves by 2\%, i.e., achieving 89.71, which is impressively high. Also, on TREC-QA, ASR outperforms all models, being on par with PR regarding P@1. The latter is 97.06, which corresponds to mistaking the answers of only two questions. We manually checked these and found out that these were two annotation errors: ASR achieves perfect accuracy while PR only mistakes one answer. Of course, this just provides evidence that PR based on RoBERTa-Large solves the task of selecting the best answers (i.e., measuring P@1 on this dataset is not meaningful anymore).

\begin{table}[t]
\centering
\resizebox{1\linewidth}{!}{
\begin{tabular}{|p{5.5em}|l|r|r|r|r|r|r|} 
\hline
RoBERTa Large                   	& \multicolumn{3}{c|}{WikiQA}                                                    & \multicolumn{3}{c|}{TREC-QA}                                                     \\ 
\hline
                                 			& \multicolumn{1}{l|}{P@1} & \multicolumn{1}{l|}{MAP} & \multicolumn{1}{l|}{MRR} & \multicolumn{1}{l|}{P@1} & \multicolumn{1}{l|}{MAP} & \multicolumn{1}{l|}{MRR}  \\ 
\hline
Garg et al.,              			&           --               & 0.9200                   & 0.9330                    &             --             & 0.9430                    & 0.9740                     \\ 
\hline
Our Reranker                         		& 0.8724                   & 0.9151                   & 0.9266                   & \textbf{0.9706}                   & 0.9481                   & \textbf{0.9816}                    \\ 
\hline
KGAT (K=2) 				& 0.8642    &0.9094                   & 0.9218                   & 0.9559                   & 0.9407                   & 0.9743                    \\ 
\hline
ASR (K=3) 				& \textbf{0.8971}                 & \textbf{0.9280}                   & \textbf{0.9399}                   & \textbf{0.9706}                  &\textbf{0.9488}                   & \textbf{0.9816}                    \\ 
\hline
\end{tabular}
}
\arrayrulecolor{black}
\caption{Results on WikiQA and TREC-QA, using RoBERTa Large Transformer.}
\label{table:resultsForall_large}
\vspace{-1em}
\end{table}

\subsection{Model Discussion}
Table \ref{table:ascaccuracy} reports the accuracy of ASC inside different models. In ASR, it uses 4-way categories, while in MASR-based models, it uses the three FEVER labels (see Sec.~\ref{sec:new_model}). ACC is the overall accuracy while F1 refers to only the category 0. We note that ASC in MASR-FP achieves the highest accuracy averaged over all datasets. This happens since we pre-fine-tuned it with the FEVER data.

We analyzed examples for which ASR is correct and PR is not. Tab.~\ref{example-1} shows that, given $q$ and $k=3$ candidates, PR chooses $c_1$, a suitable but wrong answer. This probably happens since the answer best matches the syntactic/semantic patterns of the question, which ask for a \emph{type of color}, indeed, the answer offers such type, \emph{primary colors}.  PR does not rely on any background information that can support the set of colors in the answer. In contrast, ASR selects $c_2$ as it can rely on the support of other answers. Its ASC provides an average score for the category $0$ (both members are correct) of $c_2$, i.e., $\frac{1}{k}\sum_{i\neq2} \text{ASC}(c_2,c_i) = 0.653$, while for $c_1$ the average score is significant lower, i.e., 0.522. This provides higher support for $c_2$, which is used by ASR to rerank the output of PR.

\begin{table}[t]
\centering
\resizebox{.8\linewidth}{!}{
\begin{tabular}{|l|r|r|r|r|r|r|} 
\hline
        & \multicolumn{2}{c|}{WikiQA}                        & \multicolumn{2}{c|}{TREC-QA}                        & \multicolumn{2}{c|}{WQA}                           \\ 
\hline
        & \multicolumn{1}{l|}{ACC} & \multicolumn{1}{l|}{F1} & \multicolumn{1}{l|}{ACC} & \multicolumn{1}{l|}{F1} & \multicolumn{1}{l|}{ACC} & \multicolumn{1}{l|}{F1}  \\ 
\hline
ASR     & 0.59                     & 0.00                    & 0.56                     & 0.80                    & 0.58                     & 0.64                     \\ 
\hline
MASR    & 0.46                     & 0.00                    & 0.45                     & 0.62                    & 0.53                     & 0.61                     \\ 
\hline
MASR-F  & 0.46                     & 0.00                    & 0.64                     & 0.78                    & 0.58                     & 0.68                     \\ 
\hline
MASR-FP & 0.49                     & 0.37                    & 0.65                     & 0.73                    & 0.59                     & 0.69                     \\
\hline
\end{tabular}
}
\caption{The Accuracy and F1 of category 0 for ASC}
\label{table:ascaccuracy}
\vspace{-1em}
\end{table}

Tab.~\ref{example-2} shows an interesting case where all the sentences contain the required information, i.e., February. However, PR and ASR both choose answer $c_0$, which is correct but not natural, as it provides the requested information indirectly. Also, it contains a lot of ancillary information. In contrast, MASR is able to rerank the best answer, $c_1$, in the top position.

\section{Conclusion}
\vspace{-.5em}
\label{sec:con}

We have proposed new joint models for AS2.
ASR encodes the relation between the target answer and all the other candidates, using an additional Transformer model, and an Answer Support Classifier, while MASR jointly models the ASR representations for all target answers. We extensively tested KGAT, ASR, MASR, and other joint model baselines we designed.

\begin{table}[t]
\small
\centering
\resizebox{\linewidth}{!}{%
\begin{tabular}{|lp{7.2cm}|}
\hline
$q$: & \textbf{What kind of colors are in the rainbow?}\\
$c_1$: & Red, yellow, and blue are called the primary colors.\\
$c_2$: &\vspace{-0.7em} {\color{mygreen}The order of the colors in the rainbow goes: red, orange, yellow, green, blue, indigo and violet.}\\
$c_3$: &\vspace{-0.7em} {\color{mygreen}The colors in all rainbows are present in the same order: red, orange, yellow, green, blue, indigo, and violet.}\\
$c_4$: &\vspace{-0.7em} {\color{mygreen}A rainbow occurs when white light bends and separates into red, orange, yellow, green blue, indigo and violet.}\\
\hline
\end{tabular}
}
\caption{A question with answer candidates ranked by PR; ASR chose $c_2$.}
\label{example-1}
\end{table}

\begin{table}[t]
\small
\centering
\resizebox{\linewidth}{!}{%
\begin{tabular}{|lp{7.2cm}|}
\hline
$q$: & \textbf{What's the month of Valentine's day?}\\
$c_0$: & Celebrated on February 14 every year, saint Valentine's day or Valentine's day is the traditional day on which lovers convey their love to each other by sending Valentine's cards, sometimes even anonymously.\\
$c_1$: &  \vspace{-0.7em}{\color{mygreen}February is historically chosen to be the month of love and romance and the month to celebrate Valentine's day.}\\
$c_2$: & In order for today to be Valentine's day, it's necessary that today is in the month of February.\\
$c_3$: & \vspace{-0.7em}{Every year, Valentine's day is celebrated on February 14 in many countries around the world.}\\
\hline
\end{tabular}
}
\caption{A question with answer candidates $\{c_0, {\color{mygreen}c_1}, c_2, {c_3}\}$ ranked by PR; ASR reranks as $\{c_0, {c_3}, c_2, {\color{mygreen}c_1}\}$; and MASR reranks as $\{{\color{mygreen}c_1}, {c_3}, c_0, c_2\}$; ${\color{mygreen}c_1}$ is the natural correct answer.}
\label{example-2}
\end{table}

The results show that our models can outperform the state of the art. Most interestingly, ASR constantly outperforms all the models (but MASR-FP), on all datasets, through all measures, and for both base and large transformers. For example, ASR achieves the best reported results, i.e., MAP values of 92.80\% and 94.88, on \mbox{WikiQA} and TREC-QA, respectively. MASR improves ASR by  2\% on WQA, since this contains enough data to train the ASR representations jointly.

\bibliography{main}
\bibliographystyle{acl_natbib}

\end{document}